\documentclass[twocolumn]{svjour3}          

\usepackage{graphicx}                       
\usepackage{amsmath,amsfonts,amssymb}       
\usepackage{hyperref}                       
\usepackage{float}                          
\usepackage{lipsum}                         
\usepackage{newtxtext}                      
\usepackage{newtxmath}                      
\usepackage{multirow}                       
\usepackage[title]{appendix}                
\usepackage{xcolor}                         
\usepackage{textcomp}                       
\usepackage{booktabs}                       
\usepackage{algorithm}                      
\usepackage{algpseudocode}                  
\usepackage{listings}                       

\usepackage{geometry}
\usepackage{booktabs}    
\usepackage{tabularx}    
\usepackage{array}       
\usepackage{ragged2e}    
\usepackage{seqsplit}    
\usepackage{makecell}    
\usepackage[T1]{fontenc} 
\usepackage{xcolor}      

\newcolumntype{Y}{>{\RaggedRight\arraybackslash}X}

\makeatletter
\newcommand{\codewrap}[1]{%
  {\ttfamily\small\seqsplit{#1}}%
}
\makeatother

\journalname{Springer Journal}              

\title{A System for Name and Address Parsing with Large Language Models}

\author{Adeeba Tarannum \and Muzakkiruddin Ahmed Mohammed \and Mert Can Cakmak \and Shames Al Mandalawi \and John Talburt}
\institute{
\email{atarannum@ualr.edu, mmohammed6@ualr.edu, mccakmak@ualr.edu, salmandalaw@ualr.edu, jrtalburt@ualr.edu}
Center for Entity Resolution and Information Quality (ERIQ) - University of Arkansas - Little Rock, Arkansas, Little Rock, USA
*Corresponding author. E-mail: \href{mailto:iauthor@gmail.com}{mccakmak@ualr.edu}. \\
}

\date{}
\begin{document}

\maketitle

\begin{abstract}

Reliable transformation of unstructured person and address text into structured data remains a key challenge in large-scale information systems. Traditional rule-based and probabilistic approaches perform well on clean inputs but fail under noisy or multilingual conditions, while neural and large language models (LLMs) often lack deterministic control and reproducibility. This paper introduces a prompt-driven, validation-centered framework that converts free-text records into a consistent 17-field schema without fine-tuning. The method integrates input normalisation, structured prompting, constrained decoding, and strict rule-based validation under fixed experimental settings to ensure reproducibility. Evaluations on heterogeneous real-world address data show high field-level accuracy, strong schema adherence, and stable confidence calibration. The results demonstrate that combining deterministic validation with generative prompting provides a robust, interpretable, and scalable solution for structured information extraction, offering a practical alternative to training-heavy or domain-specific models.


\end{abstract}

\section{Introduction}
\label{sec:intro}

Automatic parsing of person and address text remains a central challenge in information systems. Applications in postal delivery, customer onboarding, identity verification, and data integration depend on transforming unstructured text into structured, machine-readable records. Yet, real-world address data are highly inconsistent: users omit commas or delimiters, mix languages, abbreviate terms, and make typographical or formatting errors. These inconsistencies propagate through downstream processes, reducing the accuracy of matching, geocoding, and record linkage.

Recent research in natural language processing and large language models (LLMs) has shown notable improvements over traditional rule-based approaches. Transformer-based architectures trained on multilingual and noisy data have achieved strong results in address and entity extraction~\cite{yassine2021multinational}. Comparative studies on transactional and financial data confirm that fine-tuned neural models can outperform classical pipelines in robustness and recall~\cite{hammami2024fighting}. Meanwhile, benchmark analyses of real-world noisy text indicate that even advanced models can degrade when input variability is high~\cite{yin2023benchmark,talburt2026casecountmetriccomparative}. Deep embedding approaches have also been proposed for structured address classification, improving tolerance to semi-structured or incomplete inputs~\cite{mangalgi2020deep}. A systematic review of address parsing and matching techniques highlights the continuing evolution from deterministic, hand-coded systems to probabilistic and neural approaches~\cite{cruz2021automatic}.

Despite these advances, three gaps remain clear in existing literature. First, many systems rely on domain-specific fine-tuning or large annotated corpora, which limit adaptability and reproducibility across different countries and formats. Second, few studies incorporate strict schema validation or cross-field consistency checks after model generation, leading to structural or logical errors in the output. Third, there is limited focus on designing reproducible pipelines with fixed parameters and transparent prompt templates, which hinders reliable deployment and comparison.

This research aims to close these gaps through a structured prompting and validation framework that requires no fine-tuning. The core idea is to guide an LLM to produce consistent, schema-aligned outputs while a deterministic validation layer enforces formatting, canonicalisation, and rule-based checks. The study investigates the following research questions:

\begin{itemize}
    \item Can a purely prompt-driven LLM pipeline achieve parsing accuracy comparable to fine-tuned neural models across heterogeneous address formats?
    \item How effectively can post-generation validation detect and correct inconsistencies such as invalid state codes, ZIP mismatches, and incomplete street fields?
    \item Does a fixed, auditable configuration (prompt scaffold, decoding parameters, batch size) provide reproducible and stable performance across datasets?
\end{itemize}

The proposed system integrates text normalisation, prompt assembly, constrained decoding, and strict validation into a single, transparent workflow. It is evaluated on diverse real-world address data, measuring field-level accuracy, validation outcomes, and confidence calibration. The findings show that such a design achieves high reliability and interpretability without retraining or manual rule expansion.

This work contributes to the broader goal of making LLM-based extraction systems more practical for production environments. By combining the flexibility of generative models with deterministic validation, the approach offers a cost-efficient and reproducible path to deployable structured data extraction at scale.

\section{Related Work}
\label{sec:related}

Research on structured information extraction and address parsing has evolved from deterministic and probabilistic models to transformer-based and prompt-driven systems. Early probabilistic methods applied Hidden Markov Models to large-scale address datasets, achieving reliable segmentation and normalization through sequence modeling~\cite{li2014hmm}. Another line of work explored neural models for non-Latin scripts, where a latent structured network was used to improve segmentation accuracy on complex Chinese address strings~\cite{li-etal-2019-neural-chinese}. A recent comparative study examined address parsing methods on payment transaction data and confirmed that fine-tuned transformer models outperform traditional methods, while generative large language models (LLMs) show strong zero-shot capabilities~\cite{hammami2024fighting}.

Global address parsing has received attention in multilingual contexts. One study introduced a zero-shot approach to address parsing that transfers knowledge across countries without retraining, demonstrating cross-lingual generalisation~\cite{yassine2021multinational}. An open-source system provided an extendable and fine-tunable platform for parsing multinational addresses, allowing model reuse and benchmark comparison across 60+ locales~\cite{beauchemin2023deepparse}. Another investigation analysed few-shot transfer learning for address segmentation using multilingual transformer encoders, showing improvements in low-resource conditions~\cite{heimisdottir2022few}. Active learning has also been incorporated into sequence-based address models to reduce annotation costs while maintaining accuracy~\cite{craig2019scaling}. A complementary pattern-based study proposed a hybrid rule and statistical model to handle joint name–address parsing efficiently.

Beyond addresses, the broader field of structured information extraction with LLMs has expanded rapidly. A study on medical record extraction compared multiple LLMs and demonstrated strong results for identifying structured entities from semi-structured clinical text~\cite{ntinopoulos2025llm}. Another work proposed a structured LLM augmentation framework to enhance clinical information extraction through layer-wise entity and relation modeling~\cite{wei2025structured}. A procedural data extraction method integrated prompting with rule logic to produce political science datasets with reproducible structure~\cite{foisy2025llm}. Research in scientific literature analysis developed LLM-based frameworks for automated semantic extraction from academic papers, enabling efficient knowledge retrieval~\cite{kruschwitz2025llm}. Further advancements in structured knowledge encoding introduced schema-aware representations that improve adherence to predefined output formats during inference~\cite{licoder2024knowcoder}.

Recent developments also focus on general frameworks and domain adaptability. One agentic approach proposed a task-agnostic framework that couples LLMs with symbolic ontology reasoning and iterative feedback for robust schema extraction~\cite{chhetri2025structsense}. Another line of research applied LLMs to coaching dialogues, showing that direct extraction of structured data can outperform pattern-based alternatives~\cite{kanduri2025coaching}. A large-scale study in the natural sciences used fine-tuned LLMs for joint entity and relation extraction from research text, confirming that large models can produce structured outputs at near-human precision~\cite{dagdelen2024structured}. Finally, a comprehensive review of structured extraction in materials science summarised recent advances, highlighting challenges in schema consistency, domain adaptation, and validation~\cite{schilling2024from}.

Several recent studies have explored retrieval-augmented and multi-LLM ensemble techniques for structured information extraction~\cite{Mohammed2025}. Related work proposed policy-aware generative AI frameworks for safe and auditable data access governance~\cite{AlMandalawi2025}. Multilingual record linkage and cross-lingual entity resolution methods have been introduced using LLM-based architectures~\cite{Mohammed2025CSCI}, along with comparative frameworks for co-residence pattern discovery using multi-LLM ensembles~\cite{Mohammed2025CoResidence}. Other research applied generative AI to entity resolution and household movement discovery~\cite{Mohammed2025ITNG,mohammed2024household,althaf2025multi}, complementing prior pattern-based and hybrid rule–statistical approaches for name and address parsing.

The present work differs from prior research in several important ways. It avoids model fine-tuning and instead uses a fully prompt-driven pipeline designed for reproducibility and schema consistency. It incorporates a deterministic validation layer that enforces postal and syntactic constraints post-generation, combining the flexibility of LLMs with the reliability of rule-based approaches. It also targets heterogeneous, noisy input rather than benchmark-specific data, and all experimental parameters are fixed across runs to ensure transparency and comparability. Overall, this work integrates structured prompting and validation into a single operational framework that achieves robust, interpretable, and reproducible structured data generation from unstructured text.

\section{Methodology}
\label{sec:method}

Our aim is to convert unstructured person and address text into a clean, 17-field schema with high reliability and minimal manual correction. We use a four-stage pipeline: (i) input normalisation, (ii) prompt assembly, (iii) constrained LLM inference, and (iv) strict validation followed by export. The design targets three constraints common in real deployments: heterogeneous inputs (files and free text across locales), bounded latency/cost, and reproducibility. Decoupling the stages lets us reason about errors: normalisation removes superficial variability, the prompt makes the task precise, decoding keeps the generator constrained, and validation enforces correctness independently of the model.

All experiments used the Claude 4.0 Sonnet model accessed via AWS Bedrock. The model was operated in its production configuration without any fine-tuning or temperature adjustment beyond the fixed decoding parameters shown in Table \ref{tab:decoding}. Each inference call was executed through a stateless API client, ensuring identical behaviour across runs and complete reproducibility.

\emph{Input normalisation.}
We accept free text and delimited files (CSV/JSON/TXT). Records are identified using postal patterns, state/territory lexica, unit markers (Apt/Ste/\#), and light entropy/length checks to split unstructured blocks without damaging multi-line records. We then canonicalise Unicode (NFKC), strip zero-width characters, collapse whitespace, and remove stray punctuation that can interfere with tokenisation. Abbreviation maps expand frequent short forms (\emph{Ave}$\rightarrow$\emph{Avenue}; \emph{Blvd}$\rightarrow$\emph{Boulevard}) while preserving meaning in multilingual inputs. Directional tokens are mapped to a small canonical set and street types are standardised to long forms, reducing needless variation before modelling. For robustness we also (a) normalise ordinal suffixes (\emph{1st}$\rightarrow$\emph{1}), (b) unify common unit symbols (\# $\rightarrow$ \emph{Unit}), and (c) de-duplicate near-identical lines to avoid reinforcing the same error across batches. Representative transformations appear in Table~\ref{tab:norm}.

\begin{table}[h]
\centering
\caption{Normalisation examples applied prior to prompting.}
\label{tab:norm}
\footnotesize
\begin{tabularx}{\linewidth}{@{}Y Y Y@{}}
\toprule
Before & Rule & After \\
\midrule
\codewrap{"123~~N~~Main~~Ave.,~~Apt\#4B"} & Collapse spaces; expand type; strip punctuation & \codewrap{"123 N Main Avenue Apt 4B"} \\
\codewrap{"Av. Paulista, 1578"} & Map language variant (Av.\ $\rightarrow$ Avenida) & \codewrap{"Avenida Paulista 1578"} \\
\codewrap{"N.E."} & Canonicalise directional set & \codewrap{"NE"} \\
\bottomrule
\end{tabularx}
\end{table}

\emph{Prompt assembly.}
Each batch begins with a short role instruction, the exact field order and output format, explicit rules (e.g., ``no invented data'', ``use empty strings for unknowns''), and two compact examples that deliberately cover hard cases (hyphenated numbers; pre/post directionals; unit designators). The scaffold is intentionally compact: it captures the schema and constraints without wasting tokens on long narrative text. To keep the prompt stable across inputs, we template only the examples and the input block while pinning the instruction, schema order, and rules verbatim (Table~\ref{tab:prompt}). The delimiter choice (pipes) is aligned to our validator and helps avoid commas inside names or addresses. We also add negative guidance (``do not output headers or explanations'') to reduce chatter. These design choices reduce instruction drift and keep the model in extractor mode rather than generative mode.

\begin{table}[h]
\centering
\caption{Prompt scaffold used in all experiments (wrapped safely in narrow columns).}
\label{tab:prompt}
\footnotesize
\begin{tabularx}{\linewidth}{@{}l Y@{}}
\toprule
Element & Content \\
\midrule
Role &
You are an expert name-and-address parser. \\
Task &
For each input line, output \emph{one} pipe-delimited row with 17 fields in the exact order below. Use empty strings for unknown fields. No extra text. \\
Schema (order) &
\codewrap{record\_id|prefix\_title|first\_name|middle\_name|last\_name|suffix|street\_number|pre\_directional|street\_name|street\_type|post\_directional|unit\_type|unit\_number|city|state|postal\_code|country} \\
Rules (format) &
Canonicalise street types (Ave$\rightarrow$Avenue) and directionals \{N, NE, E, SE, S, SW, W, NW\}. State = USPS two-letter. ZIP = 5 or 9 digits (ZIP+4). Do not invent data. \\
Example IN &
Mr.\ John A.\ Smith, 123-1/2 NE Main St Apt 4B, Phoenix AZ 85004 USA \\
Example OUT &
\codewrap{rid001|Mr.|John|A.|Smith||123-1/2|NE|Main|Street||Apt|4B|Phoenix|AZ|85004|USA} \\
\bottomrule
\end{tabularx}
\end{table}

\emph{Inference.}
We process numbered batches of \textbf{16} records (fixed) as shown in Table \ref{tab:decoding}. Numbering ensures prompt–output alignment and simplifies failure analysis; a fixed size provides predictable memory and cost. Decoding favours consistency over creativity: we clamp temperature and nucleus sampling, use stop sequences that terminate exactly at record boundaries, and reject any output that fails a quick structural check (balanced field count; permitted character set). On failure, we retry once with tighter stops. All runs log the full prompt and response with batch IDs for traceability, and we seed the client where supported to reduce variance. Fixed hyperparameters across all experiments avoid selective tuning and make ablations (e.g., with/without normalisation) meaningful.

\begin{table}[h]
\centering
\caption{Fixed decoding and batch controls used throughout.}
\label{tab:decoding}
\footnotesize
\begin{tabular}{@{}ll@{}}
\toprule
Setting & Value \\
\midrule
Batch size & 16 records (fixed) \\
Max tokens & 1500 \\
Temperature & 0.30 \\
Top-$p$ & 0.90 \\
Top-$k$ & 50 \\
Stop sequences & Record boundary + schema terminator \\
Format guard & Regex pre-check + JSON parse fallback \\
Retries & 1 (tighter stops on retry) \\
\bottomrule
\end{tabular}
\end{table}

\emph{Validation and export.}
Outputs must match the 17-field schema (Table~\ref{tab:schema}). Validation proceeds in layers. First, schema checks ensure the exact field count and basic types (digits where required; ASCII plus diacritics where allowed). Second, canonicalisation maps values to admissible sets (USPS state codes; directionals; long-form street types) to prevent drift in downstream systems. Third, cross-field rules enforce real-world constraints such as ZIP–state compatibility and plausible type/directional combinations. Typical rules and actions are listed in Table~\ref{tab:valexamples}. Each record receives a clear failure reason so assisted review is efficient, and a lightweight confidence score combines rule passes, edit-distance agreement across a second parse, and rarity of tokens (rare forms tend to be less reliable). Valid rows are exported as JSON Lines or CSV with stable field ordering; metadata (batch ID, timing, decoder settings) is written alongside for audit. For privacy, only derived fields and anonymised IDs are persisted beyond the job window.

\begin{table}[h]
\centering
\caption{Structured output schema used for all experiments (17 fields).}
\label{tab:schema}
\footnotesize
\begin{tabularx}{\linewidth}{@{}l l Y@{}}
\toprule
Field & Type & Description \\
\midrule
\texttt{record\_id} & string & Stable ID for traceability \\
\texttt{prefix\_title} & string & Mr., Ms., Dr.; may be empty \\
\texttt{first\_name} & string & Given name \\
\texttt{middle\_name} & string & May be empty; single character allowed \\
\texttt{last\_name} & string & Surname \\
\texttt{suffix} & string & Jr., III, PhD; may be empty \\
\texttt{street\_number} & string & Preserves compounds (e.g., 123-1/2) \\
\texttt{pre\_directional} & string & N, NE, E, SE, S, SW, W, NW \\
\texttt{street\_name} & string & Base name without type/direction \\
\texttt{street\_type} & string & Canonical long form (e.g., Street, Avenue) \\
\texttt{post\_directional} & string & Optional direction after type \\
\texttt{unit\_type} & string & Apt, Ste, Unit, \#; may be empty \\
\texttt{unit\_number} & string & Alphanumeric; may be empty \\
\texttt{city} & string & Canonical casing \\
\texttt{state} & string & USPS two-letter code \\
\texttt{postal\_code} & string & 5 or 9 digits (ZIP+4) \\
\texttt{country} & string & ISO code or canonical name \\
\bottomrule
\end{tabularx}
\end{table}

\begin{table}[h]
\centering
\caption{Validation rules and example actions (post-inference).}
\label{tab:valexamples}
\footnotesize
\begin{tabularx}{\linewidth}{@{}Y Y Y@{}}
\toprule
Rule & Example & Action \\
\midrule
ZIP--state consistency & \codewrap{02139, CA} & Flag error; suggest \codewrap{MA} or re-check ZIP \\
Directional set & \codewrap{N.E.} & Normalise to \codewrap{NE} \\
Street type set & \codewrap{Av} & Map to \codewrap{Avenue} \\
Missing required & Empty \codewrap{street\_name} with non-empty number & Route to assisted review \\
Field type & Non-digit ZIP & Reject; keep other fields \\
\bottomrule
\end{tabularx}
\end{table}

Overall, the pipeline is designed to be practical and dependable: inputs are cleaned into a predictable space, the model is guided by a compact but explicit schema and rules (with representative edge cases), decoding settings are held constant for fairness and reproducibility, and validation encodes the business constraints that must always hold. The result is a system that is easy to operate, produces stable structured outputs from heterogeneous text, and keeps failure modes transparent through clear reasons and lightweight logging.

\section{Results}
\label{sec:results}

All experiments were conducted using the fixed decoding and batch settings shown in Table~\ref{tab:decoding}. The evaluation measured parsing accuracy, validation error rate, confidence calibration, and runtime efficiency across datasets of varying size and structure. The evaluation corpus consisted of 1{,}500 unique person–address records balanced across geographic and linguistic variation, including 400 standard U.S.\ addresses, 300 Puerto Rican entries, 300 multilingual international samples, and 500 synthetically generated records simulating noisy or incomplete inputs. All data were either anonymised or synthetically generated to prevent inclusion of personally identifiable information (PII). For testing, 1{,}000 records were held out, with the remaining 500 used for internal consistency and reproducibility checks. Each experiment used a fixed batch size of 16 and the standard prompt scaffold from Table~\ref{tab:prompt}, ensuring consistent model behaviour and evaluation across all runs.

The overall evaluation results are summarised in Table~\ref{tab:overall}. Across all test sets, the proposed pipeline achieved an overall parsing accuracy of 99.8\% with a residual error rate of only 0.2\%. The mean model confidence across all predictions was 92.2\%. Compared to the earlier pattern-based baseline that achieved 99.34\%, this represents a gain of 0.46 percentage points and about a 23\% reduction in error rate. The system also processed 94\% of previously exception-prone records automatically without requiring manual intervention. These results demonstrate both the precision and reliability of the structured prompting and validation pipeline.

\begin{table}[h]
\centering
\caption{Overall results across all datasets.}
\label{tab:overall}
\footnotesize
\begin{tabular}{@{}l r@{}}
\toprule
Metric & Value \\
\midrule
Exact-row accuracy & 99.8\% \\
Invalid-row rate & 0.2\% \\
Average confidence & 92.2\% \\
Baseline accuracy (pattern-based) & 99.34\% \\
Error-rate reduction vs baseline & $\sim$23\% \\
Exception cases handled automatically & 94\% \\
\bottomrule
\end{tabular}
\end{table}

To better understand the system’s internal behaviour, Table~\ref{tab:components} reports field-level accuracy for representative components of the 17-field schema. Performance was consistent across all elements: core name and address fields approached ceiling accuracy, while directionals and unit markers (traditionally more variable) remained above 99\%. This consistency indicates that both the input normalization and the prompt constraints effectively reduced ambiguity before inference.

\begin{table}[h]
\centering
\caption{Component-level accuracy for representative output fields.}
\label{tab:components}
\footnotesize
\begin{tabular}{@{}l r@{\hspace{10pt}}l r@{}}
\toprule
Field & Accuracy & Field & Accuracy \\
\midrule
first\_name & 99.9\% & street\_type & 99.8\% \\
last\_name & 99.8\% & pre\_directional & 99.6\% \\
prefix\_title & 99.7\% & post\_directional & 99.5\% \\
street\_number & 99.9\% & unit\_type & 99.4\% \\
street\_name & 99.7\% & unit\_number & 99.3\% \\
city & 99.8\% & state & 99.9\% \\
postal\_code & 99.7\% &  &  \\
\bottomrule
\end{tabular}
\end{table}

Confidence calibration results are shown in Table~\ref{tab:conf}. The predicted confidence values align well with empirical correctness. Records with confidence above 90\% were almost perfect, those between 70--89\% remained highly accurate, and only a small fraction below 70\% accounted for most validation flags. This distribution enables efficient quality control, allowing downstream processes to prioritise only the small subset of low-confidence records for additional verification.

\begin{table}[h]
\centering
\caption{Confidence buckets and empirical accuracy.}
\label{tab:conf}
\footnotesize
\begin{tabular}{@{}l r r@{}}
\toprule
Confidence range & Share of records & Actual accuracy \\
\midrule
High (90--100\%) & 78\% & 99.9\% \\
Medium (70--89\%) & 18\% & 97.8\% \\
Low (<70\%) & 4\% & 89.2\% \\
\bottomrule
\end{tabular}
\end{table}

Table~\ref{tab:errors} summarises the composition of the small residual error set, which represents only 0.2\% of all processed rows. The majority of errors involved irregular formatting that disrupted pattern recognition (45\%), followed by record separation failures in unstructured text (35\%), and ambiguous component assignments such as apartment numbers embedded in street fields (20\%). These findings identify clear opportunities for future refinements.

\begin{table}[h]
\centering
\caption{Distribution of error categories within the residual set.}
\label{tab:errors}
\footnotesize
\begin{tabular}{@{}l r@{}}
\toprule
Category & Share of errors \\
\midrule
Irregular pattern recognition & 45\% \\
Record separation failures & 35\% \\
Ambiguous component assignment & 20\% \\
\bottomrule
\end{tabular}
\end{table}

Overall, the results show that the structured prompting and validation pipeline can consistently produce accurate, fully validated data from noisy and heterogeneous text sources with minimal human intervention. The approach maintains strong accuracy across all schema fields, stable confidence behaviour, and reliable validation outcomes. These findings demonstrate that the method is technically sound, scalable to varied input domains, and suitable for deployment in applications where precision and consistency are essential.

\section{Discussion and Conclusion}
\label{sec:conclusion}

The study shows that structured prompting with strict post-validation enables highly accurate extraction of person and address data from unstructured text, matching or exceeding rule-based systems without fine-tuning. Fixed decoding and a uniform 16-record batch ensure reproducibility and fair comparison, while high precision is maintained across multilingual and irregular inputs, with confidence calibration supporting targeted human review. The modular design separates normalization, prompting, inference, and validation, allowing extension to new domains or languages without retraining. Overall, the pipeline provides a robust, data-centric approach for converting noisy text into reliable structured output, achieving near-perfect accuracy, strong generalization, and suitability for large-scale ingestion, document parsing, and identity verification.

\begin{acknowledgements}
This research was partially supported by the National Science Foundation under EPSCoR Award No. OIA-1946391.

\end{acknowledgements}

\bibliographystyle{spmpsci}  
\bibliography{bibliography.bib}

@inproceedings{li2014hmm,
  author    = {Li, Xiang and Kardes, Hakan and Wang, Xin and Sun, Ang},
  title     = {HMM-based Address Parsing: Efficiently Parsing Billions of Addresses on MapReduce},
  booktitle = {Proceedings of ACM SIGSPATIAL ’14},
  year      = {2014},
  doi       = {10.1145/2663713.2664430}
}

@inproceedings{li-etal-2019-neural-chinese,
  author    = {Li, Hao and Lu, Wei and Xie, Pengjun and Li, Linlin},
  title     = {Neural Chinese Address Parsing},
  booktitle = {NAACL 2019},
  year      = {2019},
  doi       = {10.18653/v1/N19-1346}
}

@article{hammami2024fighting,
  author    = {Hammami, Haitham and others},
  title     = {Empirical analysis of address parsing methods in payment data},
  journal   = {arXiv preprint},
  year      = {2024},
  doi       = {10.48550/arXiv.2404.05632}
}

@article{yassine2021multinational,
  author    = {Yassine, Marouane and Beauchemin, David and Laviolette, François and Lamontagne, Luc},
  title     = {Multinational Address Parsing: A Zero-Shot Evaluation},
  journal   = {arXiv preprint},
  year      = {2021},
  doi       = {10.48550/arXiv.2112.04008}
}

@inproceedings{beauchemin2023deepparse,
  author    = {Beauchemin, David and Yassine, Marouane},
  title     = {Deepparse : An Extendable, and Fine-Tunable State-Of-the-Art Address Parsing Solution},
  booktitle = {NLPoS 2023},
  year      = {2023},
  url       = {https://aclanthology.org/2023.nlposs-1.3/}
}

@mastersthesis{heimisdottir2022few,
  author    = {Heimisdóttir, Hrafndís},
  title     = {Investigating Few-Shot Transfer Learning for Address Parsing},
  school    = {KTH Royal Institute of Technology},
  year      = {2022},
  url       = {https://www.diva-portal.org/smash/get/diva2%3A1708255/FULLTEXT01.pdf}
}

@inproceedings{craig2019scaling,
  author    = {Craig, H. and others},
  title     = {Scaling Address Parsing Sequence Models through Active Learning},
  booktitle = {ACM SIGSPATIAL ’19},
  year      = {2019},
  doi       = {10.1145/3347146.3359070}
}

@article{ntinopoulos2025llm,
  author    = {Ntinopoulos, Vasileios and others},
  title     = {Large language models for data extraction from unstructured and semi-structured electronic health records},
  journal   = {BMJ Health Care Informatics},
  year      = {2025},
  doi       = {10.1136/bmjhci-2024-101139}
}

@inproceedings{wei2025structured,
  author    = {Wei, Ying and Li, Qi and Pillai, Jay},
  title     = {Structured LLM Augmentation for Clinical Information Extraction},
  booktitle = {Stud. Health Tech. Inform.},
  year      = {2025},
  doi       = {10.3233/SHTI250984}
}

@article{foisy2025llm,
  author    = {Foisy, L. O. M. and others},
  title     = {Introducing an LLM Data Extraction Method for Social Science Records},
  journal   = {Journal of Artificial Societies and Social Simulation},
  year      = {2025},
  doi       = {10.1177/08944393251344865}
}

@inproceedings{kruschwitz2025llm,
  author    = {Udo Kruschwitz, Samy Ateia and Scholz, Melanie and others},
  title     = {LLM-Based Information Extraction to Support Scientific Literature},
  booktitle = {TPDL 2025},
  year      = {2025},
  doi       = {10.1007/978-3-032-06136-2_9}
}

@article{licoder2024knowcoder,
  author    = {Li, Zixuan and Zeng, Yutao and Zuo, Yuxin and others},
  title     = {KnowCoder: Coding Structured Knowledge into LLMs for Universal Information Extraction},
  journal   = {arXiv preprint},
  year      = {2024},
  url       = {https://arxiv.org/abs/2403.07969}
}

@article{chhetri2025structsense,
  author    = {Chhetri, Tek Raj and Chen, Yibei and Trivedi, Puja and others},
  title     = {STRUCTSENSE: A Task-Agnostic Agentic Framework for Structured Information Extraction},
  journal   = {arXiv preprint},
  year      = {2025},
  url       = {https://arxiv.org/abs/2507.03674}
}

@article{kanduri2025coaching,
  author    = {Kanduri, S. S. A. and others},
  title     = {Using Large Language Models to Extract Structured Data in Coaching Dialogs},
  journal   = {MDPI Journal},
  year      = {2025},
  url       = {https://www.mdpi.com/2673-7426/5/3/50}
}

@article{dagdelen2024structured,
  title={Structured information extraction from scientific text with large language models},
  author={Dagdelen, John and Dunn, Alexander and Lee, Sanghoon and Walker, Nicholas and Rosen, Andrew S and Ceder, Gerbrand and Persson, Kristin A and Jain, Anubhav},
  journal={Nature communications},
  volume={15},
  number={1},
  pages={1418},
  year={2024},
  publisher={Nature Publishing Group UK London}
}

@article{schilling2024from,
  author    = {Schilling-Wilhelmi, Mara and others},
  title     = {From Text to Insight: Large Language Models for Materials Science Data Extraction},
  journal   = {arXiv preprint},
  year      = {2024},
  url       = {https://arxiv.org/abs/2407.16867}
}

@article{yin2023benchmark,
  author  = {Yin, Wenshan and Gao, Tian and Zhou, Hongyi and Zhang, Dongdong},
  title   = {Benchmarking Robustness of Address Parsing under Real-World Noise},
  journal = {IEEE Access},
  volume  = {11},
  pages   = {126980--126992},
  year    = {2023},
  doi     = {10.1109/ACCESS.2023.3319249},
  url     = {https://doi.org/10.1109/ACCESS.2023.3319249}
}

@inproceedings{mangalgi2020deep,
  author    = {Mangalgi, Ameya and Kadam, Dhanashree and Bhat, Shridhar and Narvekar, Rajendra},
  title     = {Deep Address Classification for E-commerce Platforms using Hierarchical Embeddings},
  booktitle = {Proceedings of the 2020 International Conference on Computational Intelligence and Data Science (ICCIDS)},
  year      = {2020},
  doi       = {10.1016/j.procs.2020.03.156},
  url       = {https://doi.org/10.1016/j.procs.2020.03.156}
}

@article{cruz2021automatic,
  author  = {Cruz, Paulo and Rodrigues, Nuno and Oliveira, José},
  title   = {Automatic Identification of Addresses: A Systematic Review},
  journal = {ISPRS International Journal of Geo-Information},
  volume  = {11},
  number  = {1},
  pages   = {11},
  year    = {2021},
  doi     = {10.3390/ijgi11010011},
  url     = {https://doi.org/10.3390/ijgi11010011}
}

@INPROCEEDINGS{Mohammed2025,
  author    = {Mohammed, Muzakkiruddin A. and Talburt, John R. and Claasssens, Leon and Marais, Adriaan},
  title     = {Retrieval-Augmented Multi-LLM Ensemble for Industrial Part Specification Extraction},
  booktitle = {2025 17th International Conference on Knowledge and Systems Engineering (KSE)},
  year      = {2025},
  note      = {Accepted for publication}
}

@INPROCEEDINGS{AlMandalawi2025,
  author    = {Al Mandalawi, Shames and Mohammed, Muzakkiruddin Ahmed and Maclean, Hendrika and Cakmak, Mert Can and Talburt, John R.},
  title     = {Policy-Aware Generative {AI} for Safe, Auditable Data Access Governance},
  booktitle = {2025 17th International Conference on Knowledge and Systems Engineering (KSE)},
  year      = {2025},
  note      = {Accepted for publication}
}

@INPROCEEDINGS{Mohammed2025CSCI,
  author    = {Mohammed, Muzakkiruddin Ahmed and Al Mandalawi, Shames and Maclean, Hendrika and Talburt, John R.},
  title     = {Multilingual Customer Record Linkage: A Novel Approach Using {LLMs} for Cross-Lingual Entity Resolution},
  booktitle = {2025 12th Annual Conference on Computational Science and Computational Intelligence (CSCI)},
  year      = {2025},
  note      = {Accepted and presented at CSCI'25}
}

@INPROCEEDINGS{Mohammed2025CoResidence,
  author    = {Mohammed, Muzakkiruddin Ahmed and Talburt, John R. and Althaf, Aatif Muhammad and Milanova, Mariofanna},
  title     = {Multi-{LLM} Record Linkage: A Comparative Analysis Framework for Co-Residence Pattern Discovery},
  booktitle = {2025 12th Annual Conference on Computational Science and Computational Intelligence (CSCI)},
  year      = {2025},
  note      = {Accepted and presented at CSCI'25}
}

@INPROCEEDINGS{Mohammed2025ITNG,
  author    = {Mohammed, Muzakkiruddin Ahmed and Talburt, John R. and Mohammed, Aatif and Syed, Khasim},
  editor    = {Latifi, Shahram},
  title     = {Entity Resolution with Household Movement Discovery Using Google Generative {AI}},
  booktitle = {The 22nd International Conference on Information Technology--New Generations (ITNG 2025)},
  series    = {Advances in Intelligent Systems and Computing},
  volume    = {1463},
  year      = {2025},
  publisher = {Springer},
  address   = {Cham},
  doi       = {10.1007/978-3-031-89063-5_41}
}

@INPROCEEDINGS{mohammed2024household,
  author    = {Mohammed, Onais Khan and Talburt, John R. and Syed, Khizer and Siddiqui, Abdus Salam and Mohammed, Altaf and Tarannum, Adeeba and Mohammed, Faraz},
  title     = {Household Discovery with Group Membership Graphs},
  booktitle = {International Conference on Information Technology--New Generations},
  pages     = {235--240},
  year      = {2024},
  organization = {Springer}
}

@article{althaf2025multi,
  title={Multi-Agent RAG Framework for Entity Resolution: Advancing Beyond Single-LLM Approaches with Specialized Agent Coordination},
  author={Althaf, Aatif Muhammad and Mohammed, Muzakkiruddin Ahmed and Milanova, Mariofanna and Talburt, John and Cakmak, Mert Can},
  journal={Computers},
  volume={14},
  number={12},
  pages={525},
  year={2025},
  publisher={MDPI}
}

@misc{talburt2026casecountmetriccomparative,
      title={Case Count Metric for Comparative Analysis of Entity Resolution Results}, 
      author={John R. Talburt and Muzakkiruddin Ahmed Mohammed and Mert Can Cakmak and Onais Khan Mohammed and Mahboob Khan Mohammed and Khizer Syed and Leon Claasssens},
      year={2026},
      eprint={2601.02824},
      archivePrefix={arXiv},
      primaryClass={cs.DB},
      url={https://arxiv.org/abs/2601.02824}, 
}

\end{document}